# Distribution-Independent Evolvability of Linear Threshold Functions

Vitaly Feldman
IBM Almaden Research Center
vitaly@post.harvard.edu

March 15, 2018

**Abstract**

Valiant's (2007) model of evolvability models the evolutionary process of acquiring useful functionality as a restricted form of learning from random examples. Linear threshold functions and their various subclasses, such as conjunctions and decision lists, play a fundamental role in learning theory and hence their evolvability has been the primary focus of research on Valiant's framework (2007). One of the main open problems regarding the model is whether conjunctions are evolvable distribution-independently (Feldman and Valiant, 2008). We show that the answer is negative. Our proof is based on a new combinatorial parameter of a concept class that lower-bounds the complexity of learning from correlations.

We contrast the lower bound with a proof that linear threshold functions having a non-negligible margin on the data points are evolvable distribution-independently via a simple mutation algorithm. Our algorithm relies on a non-linear loss function being used to select the hypotheses instead of 0-1 loss in Valiant's (2007) original definition. The proof of evolvability requires that the loss function satisfies several mild conditions that are, for example, satisfied by the quadratic loss function studied in several other works (Michael, 2007; Feldman, 2009; Valiant, 2010). An important property of our evolution algorithm is monotonicity, that is the algorithm guarantees evolvability without any decreases in performance. Previously, monotone evolvability was only shown for conjunctions with quadratic loss (Feldman, 2009) or when the distribution on the domain is severely restricted (Michael, 2007; Feldman, 2009; Kanade *et al.*, 2010).

## 1 Introduction

Evolution is the source of the spectacularly complex organisms and behavior that we see around us. Yet we know very little about the computational mechanisms that can lead to such complexity while respecting the constraints of the Darwinian evolutionary process and using a plausible amount of resources. Recently Valiant suggested that an appropriate framework for understanding the power of evolution to produce complex behavior is that of computational learning theory [26] since both evolution and learning involve processes that adapt their behavior on the basis of experience. Accordingly, in his model, evolvability of a certain useful functionality is cast as a problem of learning the desired functionality through a process in which, at each step, the most "fit" candidate function is chosen from a small pool of mutations of the current candidate. Limits on the number of steps and the amount of computation performed at each step are imposed to make this process naturally plausible. A class of functions $C$ is considered evolvable if there exists a single representation scheme $R$ and a mutation algorithm $M$ on $R$ that, when guided by such selection, guarantees convergence to the desired function for every function in $C$. Here the requirements closely follow those of the celebrated PAC learning model [26]. In fact, every evolution algorithm (here and below in the sense defined in Valiant's model) can be simulated by an algorithm that is given random examples of the desired function. In addition, many properties of learning algorithms such as distribution-independence, weakness and attribute-efficiency apply equally to evolvability.

### 1.1 Prior Work

The constrained way in which evolution algorithms have to converge to the target function makes finding such algorithms a substantially more involved task than designing PAC learning algorithms. Initially, only the evolvability of monotone conjunctions of Boolean variables, and only when the distribution over the domain is uniform, was demonstrated (if not specified otherwise, the domain is $\{0, 1\}^n$) [27]. Subsequently this result was simplified [3] and strengthened to general conjunctions [13, 14]. Later Michael [22] described an algorithm for evolving decision lists over the uniform distribution that used a larger space of hypotheses and a different performance metric over hypotheses (specifically, quadratic loss). In our earlier work we showed that evolvability is, at least within polynomial limits, equivalent to learning by a natural restriction of well-studied statistical queries (SQ)[15], referred to as *correlational statistical queries* (CSQ) [4]. This result gives distribution-specific algorithms for any SQ learnable class of functions. By characterizing weak distribution-independent evolvability and using communication-complexity-based lower bounds [25, 2], we also proved that general linear threshold functions (also referred to as halfspaces) and even decision lists are not evolvable distribution-independently.

| Distribution | Concept class | Loss | Monotone | References |
|---|---|---|---|---|
| Uniform | monotone conjunctions | Boolean | yes | [27, 14] |
| Uniform | conjunctions | Boolean | yes | [13, 14] |
| any spherically symmetric | homogeneous LTFs | Boolean | yes | [14] |
| any product normal | homogeneous LTFs | Boolean | yes | [14] |
| All | Single points | Boolean | yes | [5] |
| Any family $\mathcal{D}$ | any CSQ learnable over $\mathcal{D}$ | Boolean | no | [4] |
| Fixed $D$ | any SQ learnable over $D$ | Boolean | no | [4] |
| Uniform | decision lists | quadratic | yes | [22] |
| All | conjunctions | quadratic | yes | [6] |
| Any family $\mathcal{D}$ | any SQ learnable over $\mathcal{D}$ | quadratic | no | [5] |
| Fixed $D$ | any SQ learnable over $D$ | quadratic | yes | [6] |

Table 1: Positive results on evolvability. For the distribution entry "All" refers to distribution-independent evolvability. $\mathcal{D}$ refers to any fixed set of distribution (including "All"). All results for Boolean loss also apply to all other loss functions.

In another work [5] we examined the relative power of a number of variants of the model discussed in Valiant's and other works [27]. Among them we considered a generalization of the model to real-valued hypotheses for which one needs to specify the loss function used to measure the loss in performance at every point. We demonstrated that a number of variants of the model are all equivalent to learning by CSQs and hence to the original model [5]. The only two properties which we found to influence the power of the model are the choice of the loss function (with the original 0/1 loss being equivalent to evolving with the linear loss) and monotonicity, or requirement that the performance of hypotheses does not decrease in the course of evolution. Valiant's original selection rule allows small decreases in performance[1]. This somewhat unnatural property has been exploited in all the results showing equivalence to learning by CSQs[2] and hence evolution algorithms obtained through such general transformations are non-monotone. In a recent work Kanade et al. [14] show that the equivalence to learning by CSQs still holds if the total allowed decrease in performance is bounded by any non-negligible value chosen in advance (they refer to such algorithms as *quasi-monotone*). The first general transformation that yields monotone algorithms was given in our subsequent work [6] where we showed that every concept class SQ learnable over a fixed distribution $D$ is evolvable monotonically over $D$ when using quadratic loss. By exploiting some of the techniques of the general transformation, we also showed that conjunctions are evolvable distribution-independently when using quadratic loss [6]. We summarize these results and several other known evolution algorithms in Table 1.

## 1.2 Our Results

As can be seen from Table 1, evolvability of even the most basic concept classes is still only partially understood. Most notably, prior to this work it was unknown whether conjunctions are evolvable distribution-independently with Boolean loss (even without requiring monotonicity) and this question was posed by Valiant and the author as an open problem at COLT 2008 [7]. In our first result (Section 3) we show that the answer is negative. Specifically, we prove that for any $k = \omega(1)$, monotone conjunctions of at most $k$ variables are not evolvable distribution-independently to any accuracy $\epsilon = o(1)$. Our technique is based on a new combinatorial parameter of a concept class that, roughly, measures the maximum number of correlational query functions required to distinguish every target function-distribution pair from a fixed function-distribution pair. This general approach is based on our recent characterization of strong SQ learnability [6]. For a given size of conjunction $k$, we then come up with a construction of a set of conjunction-distribution pairs $\{(t_S, D_S) \mid |S| = k\}$ that cannot be distinguished from a constant function over the uniform distribution using a polynomial number of queries. The distribution $D_S$ is designed in such a way that it hides all Fourier coefficients of the conjunction $t_S$ up to degree $k/3$. Simple facts from Fourier analysis of Boolean functions then imply that distinguishing between a superpolynomial number of such conjunction-distribution pairs is impossible using a polynomial number of queries.

---
[1] In this context we refer to empirical performance rather than true expected performance.
[2] The decreases in performance can be avoided if the evolution algorithm starts in a certain fixed state, i.e. is initialized.

We interpret this negative result as highlighting significant limitations of evolvability based on the Boolean feedback only. We note that many functions in biological evolution are not Boolean. For example, for most genes the amount of gene expression (that is the amount of protein produced) can vary in a certain range continuously (up to, of course, the granularity of a single molecule). Therefore it is natural to assume that when evolving the optimal regulation of gene expression (described by a Boolean function), intermediate amounts of the protein will be produced. The intermediate values are likely to cause intermediate values of loss relative to the optimal 0 or 1 value. It is therefore important to understand evolvability with other loss functions. Toward this goal in Section 4 we show that linear threshold functions are evolvable monotonically and distribution-independently for quadratic loss function and all other loss functions satisfying a set of mild conditions. We refer to loss functions that satisfy the required conditions as *well-behaved*. The amount of resources required by our algorithm depends quadratically on $1/\gamma$ where $\gamma$ is the margin of the target halfspace on the data points. Therefore, like the famous Perceptron and Winnow algorithms [23, 21], it is efficient only when the margin is non-negligible or lower bounded by the inverse of a polynomial in $n$. In the Support Vector Machine (SVM) literature this condition is usually referred to as having a *large* margin. Further, the representation used by our evolution algorithm is similar to linear thresholds and the mutation algorithm is fairly simple and natural. The only operations it requires are adding the function $\alpha \cdot x_i$ to the the current function for a real $\alpha$ and bounding the value of the function to be in $[-1, 1]$.

A very popular and powerful approach to learning when data points are not linearly separable is to embed the data points in a different (often higher dimensional) Euclidean space where the examples become linearly separable and then use a halfspace learning algorithm such as Perceptron or SVM to produce a classifier. Such approach also works in the context of evolvability and implies monotone evolvability of any concept class that can be efficiently embedded into large-margin halfspaces over some Euclidean space (efficiency of the embedding also bounds the dimension of the space). Therefore our second result approaches some of the most important and strongest results for PAC learning while also being a natural algorithm in Valiant's framework of evolvability.

We note that a similar mutation algorithm was used in our result for conjunctions [6]. However our analysis here is new and differs conceptually from the analysis for conjunctions which cannot be extended to halfspaces. It also gives substantially stronger bounds. For example, it improves the dependence of the improvement in each step on $\epsilon$ from $\epsilon^6$ to $\epsilon^2$. The key to this result for the quadratic loss function is a simple proof that for every distribution $D$, halfspace $f$ and any real-valued function $\phi$ with the range in $[-1, 1]$, there exists a variable $x_i$ that is correlated with the gradient of the loss function at point $\phi$. The absolute value of the correlation is lower-bounded by the inverse of a polynomial in $n$, $1/\epsilon$ and $1/\gamma$ and therefore is sufficient to imply that a small step in the direction of $x_i$ (or $-x_i$) will reduce the loss.

A recent work by P. Valiant [28] examines the extension of the model of evolvability to real-valued target functions. His results paint a picture quite similar to what we know about the evolvability of Boolean functions. In particular, his simple algorithm for evolving linear functions when using the quadratic loss can be seen as the counterpart of our algorithm for halfspaces.

## 2 Preliminaries

For a positive integer $\ell$, let $[\ell]$ denote the set $\{1, 2, \ldots, \ell\}$ and for $i \leq \ell$ let $[i..\ell]$ denote the set $\{i, i+1, \ldots, \ell\}$. We denote the domain of our learning problems by $X$. As usual it is parameterized by an (implicit) dimension $n$. A *concept class* over $X$ is a set of $\{-1, 1\}$-valued functions over $X$ referred to as *concepts*. Let $\mathcal{F}_1^\infty$ denote the set of all functions from $X$ to $[-1, 1]$ (that is all the functions with $L_\infty$ norm bounded by 1). It will be convenient to view a distribution $D$ over $X$ as defining the product $\langle \phi, \psi \rangle_D = \mathbf{E}_{x \sim D}[\phi(x) \cdot \psi(x)]$ over the space of real-valued functions on $X$. It is easy to see that this is simply a non-negatively weighted version of the standard dot product over $\mathbb{R}^X$ and hence is a positive semi-inner product over $\mathbb{R}^X$. The corresponding norm is defined as $\|\phi\|_D = \sqrt{\mathbf{E}_D[\phi^2(x)]} = \sqrt{\langle \phi, \phi \rangle_D}$.

Let $B_n = \{x \mid \|x_i\| \leq 1\}$ be the ball or radius 1 in $\mathbb{R}^n$, $X$ be a subset of $B_n$, and $f = \texttt{sign}(\sum w_i x_i - \theta)$ be a linear threshold function (halfspace). We define the margin $\gamma$ of $f$ on $X$ as $\gamma = \inf_{x \in X}\{|\sum w_i x_i - \theta|\}$. For convenience we use $x_0$ to refer to the constant function 1.

## 2.1 PAC Learning

The models we consider are based on the well-known PAC learning model introduced by Valiant [26]. Let $C$ be a concept class over $X$. In the basic PAC model a learning algorithm is given examples of an unknown function $f$ from $C$ on points randomly chosen from some unknown distribution $D$ over $X$ and should produce a hypothesis $h$ that approximates $f$. Formally, an *example oracle* $\text{EX}(f, D)$ is an oracle that upon being invoked returns an example $\langle x, f(x) \rangle$, where $x$ is chosen randomly with respect to $D$, independently of any previous examples.

An algorithm is said to PAC learn $C$ in time $t$ if for every $\epsilon > 0$, $f \in C$, and distribution $D$ over $X$, the algorithm given $\epsilon$ and access to $\text{EX}(f, D)$ outputs, in time $t$ and with probability at least $2/3$, a hypothesis $h$ that is evaluatable in time $t$ and satisfies $\mathbf{Pr}_D[f(x) \neq h(x)] \leq \epsilon$. We say that an algorithm *efficiently* learns $C$ when $t$ is upper bounded by a polynomial in $n$, $1/\epsilon$.

The basic PAC model is also referred to as *distribution-independent* learning to distinguish it from *distribution-specific* PAC learning in which the learning algorithm is required to learn only with respect to a single distribution $D$ known in advance. More generally, following Kearns et al. [17], one can analogously define the learnability of a set of distribution-function pairs over the same domain $X$. Namely, a set of distribution-function pairs $\mathcal{Z}$ is PAC learnable if there exists a learning algorithm that learns $f$ over $D$ (as in the definition above) for every $(D, f) \in \mathcal{Z}$.

A *weak* learning algorithm [16] is a learning algorithm that produces a hypothesis whose disagreement with the target concept is noticeably less than $1/2$ (and not necessarily any $\epsilon > 0$). More precisely, a weak learning algorithm produces a Boolean hypothesis $h$ such that $\mathbf{Pr}_D[f(x) \neq h(x)] \leq 1/2 - 1/p(n)$ for some fixed polynomial $p$.

## 2.2 The Statistical Query Learning Model

In the *statistical query model* of Kearns [15] the learning algorithm is given access to $\text{STAT}(f, D)$ – a *statistical query oracle* for target concept $f$ with respect to distribution $D$ instead of $\text{EX}(f, D)$. A query to this oracle is a function $\psi : X \times \{-1, 1\} \to \{-1, 1\}$. The oracle may respond to the query with any value $v$ satisfying $|\mathbf{E}_D[\psi(x, f(x))] - v| \leq \tau$ where $\tau \in [0, 1]$ is a real number called the *tolerance* of the query. An algorithm $\mathcal{A}$ is said to learn $C$ in time $t$ from statistical queries of tolerance $\tau$ if $\mathcal{A}$ PAC learns $C$ using $\text{STAT}(f, D)$ in place of the example oracle. In addition, each query $\psi$ made by $\mathcal{A}$ has tolerance $\tau$ and can be evaluated in time $t$.

The algorithm is said to (efficiently) SQ learn $C$ if $t$ is polynomial in $n$ and $1/\epsilon$, and $\tau$ is lower-bounded by the inverse of a polynomial in $n$ and $1/\epsilon$.

A *correlational* statistical query is a statistical query for a correlation of a function over $X$ with the target [1]. Namely the query function $\psi(x, \ell) \equiv \phi(x) \cdot \ell$ for a function $\phi \in \mathcal{F}_1^\infty$. A concept class is said to be CSQ learnable if it is learnable by a SQ algorithm that uses only CSQ queries.

## 2.3 Evolvability

We start by presenting a brief overview of the model. For a detailed description and intuition behind the various choices made in model the reader is referred to [27, 5]. The goal of the model is to specify how organisms can acquire complex mechanisms via a resource-efficient process based on random mutations and guided by performance-based selection. The mechanisms are described in terms of the multi argument functions they implement. The performance of such a mechanism is measured by evaluating the agreement of the mechanism with some "ideal" behavior function. The value of the "ideal" function on some input describes the most beneficial behavior for the condition represented by the input. The evaluation of the agreement with the "ideal" function is derived by evaluating the function on a moderate number of inputs drawn from a probability distribution over the conditions that arise. These evaluations correspond to the experiences of one or more organisms that embody the mechanism.

Random variation is modeled by the existence of an explicit algorithm that acts on some fixed representation of mechanisms and for each representation of a mechanism produces representations of mutated versions of the mechanism. The model requires that the mutation algorithm be efficiently implementable. Selection is modeled by an explicit rule that determines the probabilities with which each of the mutations of a mechanism will be chosen to "survive" based on the performance of all the mutations of the mechanism and the probabilities with which each of the mutations is produced by the mutation algorithm.

As can be seen from the above description, a performance landscape (given by a specific "ideal" function and a distribution over the domain), a mutation algorithm, and a selection rule jointly determine how each step of an evolutionary process is performed. A class of functions $C$ is considered evolvable if there exist a representation of mechanisms $R$ and a mutation algorithm $M$ such that for every "ideal" function $f \in C$, a sequence of evolutionary steps starting from any representation in $R$ and performed according to the description above "converges" in a polynomial number of steps to $f$. This process is essentially PAC learning of $C$ with the selection rule (rather than explicit examples) providing the only target-specific feedback. We now define the model formally using the notation from [5].

## 2.4 Definition of Evolvability

The description of an evolution algorithm $\mathcal{A}$ consists of the definition of the representation class $R$ of possibly randomized hypotheses in $\mathcal{F}_1^\infty$ and the description of polynomial time mutation algorithm that for every $r \in R$ and $\epsilon > 0$ outputs a random mutation of $r$.

**Definition 2.1** *A evolution algorithm $\mathcal{A}$ is defined by a pair $(R, M)$ where*

- *$R$ is a representation class of functions over $X$ with range in $[-1, 1]$.*

- *$M$ is a randomized polynomial time algorithm that, given $r \in R$ and $\epsilon$ as input, outputs a representation $r_1 \in R$ with probability $\mathbf{Pr}_\mathcal{A}(r, r_1)$. The set of representations that can be output by $M(r, \epsilon)$ is referred to as the* neighborhood *of $r$ for $\epsilon$ and denoted by $\mathtt{Neigh}_\mathcal{A}(r, \epsilon)$.*

A *loss function* $L$ on a set of values $Y$ is a non-negative mapping $L : Y \times Y \to \mathbb{R}^+$. $L(y, y')$ measures the "distance" between the desired value $y$ and the predicted value $y'$. In the context of learning Boolean functions using hypotheses with values in $[-1, 1]$ we only consider functions $L : \{-1, 1\} \times [-1, 1] \to \mathbb{R}^+$. Valiant's original model only considers Boolean hypotheses and hence only the disagreement loss (or 0-1 loss) which is equal to $L_\Delta(y, y') = y \cdot y'$. It was shown in our earlier work [5] that such loss is equivalent to the *linear loss* $L_1(y, y') = |y' - y|$ over hypotheses with the range in $[-1, 1]$. The other loss function we use here is the *quadratic loss* $L_Q(y, y') = (y' - y)^2$ function. For a function $\phi \in \mathcal{F}_1^\infty$ its performance relative to loss function $L$, distribution $D$ over the domain and target function $f$ is defined as

$$\mathtt{LPerf}_f(\phi, D) = 1 - 2 \cdot \mathbf{E}_D[L(f(x), \phi(x))]/L(-1, 1) .$$

For an integer $s$, functions $\phi, f \in \mathcal{F}_1^\infty$ over $X$, distribution $D$ over $X$ and loss function $L$, the *empirical fitness* $\mathtt{LPerf}_f(\phi, D, s)$ of $\phi$ is a random variable that equals $1 - \frac{1}{s} \frac{2}{L(-1,1)} \sum_{i \in [s]} L(f(z_i), \phi(z_i))$ for $z_1, z_2, \ldots, z_s \in X$ chosen randomly and independently according to $D$.

A number of natural ways of modeling selection were discussed in prior work [27, 5]. For concreteness here we describe the selection rule used in Valiant's main definition in a slightly generalized version from [5]. In selection rule $\mathtt{SelNB}[L, t, p, s]$ $p$ candidate mutations are sampled using the mutation algorithm. Then beneficial and neutral mutations are defined on the basis of their empirical fitness $\mathtt{LPerf}$ in $s$ experiments (or examples) using tolerance $t$. If some beneficial mutations are available one is chosen randomly according to their relative frequencies in the candidate pool. If none is available then one of the neutral mutations is output randomly according to their relative frequencies. If neither neutral or beneficial mutations are available, $\bot$ is output to mean that no mutation "survived".

**Definition 2.2** *For a loss function $L$, tolerance $t$, candidate pool size $p$, sample size $s$, selection rule $\mathtt{SelNB}[L, t, p, s]$ is an algorithm that for any function $f$, distribution $D$, mutation algorithm $\mathcal{A} = (R, M)$, a representation $r \in R$, accuracy $\epsilon$, $\mathtt{SelNB}[L, t, p, s](f, D, \mathcal{A}, r)$ outputs a random variable that takes a value $r_1$ determined as follows. First run $M(r, \epsilon)$ $p$ times and let $Z$ be the set of representations obtained. For $r' \in Z$, let $\mathbf{Pr}_Z(r')$ be the relative frequency with which $r'$ was generated among the $p$ observed representations. For each $r' \in Z \cup \{r\}$, compute an empirical value of fitness $v(r') = \mathtt{LPerf}_f(r', D, s)$. Let $\mathtt{Bene}(Z) = \{r' \mid v(r') \geq v(r) + t\}$ and $\mathtt{Neut}(Z) = \{r' \mid |v(r') - v(r)| < t\}$. Then*

(i) *if* Bene(Z) ≠ ∅ *then output* $r_1 \in$ Bene *with probability* $\mathbf{Pr}_Z(r_1)/\sum_{r' \in \text{Bene}(Z)} \mathbf{Pr}_Z(r')$;

(ii) *if* Bene(Z) = ∅ *and* Neut(Z) ≠ ∅ *then output* $r_1 \in$ Neut(Z) *with probability* $\mathbf{Pr}_Z(r_1)/\sum_{r' \in \text{Neut}(Z)} \mathbf{Pr}_Z(r')$.

(iii) *If* Neut(Z) ∪ Bene(Z) = ∅ *then output* ⊥.

A concept class $C$ is said to be evolvable by an evolution algorithm $\mathcal{A}$ guided by a selection rule Sel over distribution $D$ if for every target concept $f \in C$, mutation steps as defined by $\mathcal{A}$ and guided by Sel will converge to $f$. For simplicity here we only consider the selection rule SelNB.

**Definition 2.3** *For concept class $C$ over $X$, distribution $D$, mutation algorithm $\mathcal{A}$, loss function $L$ we say that the class $C$ is evolvable over $D$ by $\mathcal{A}$ with $L$ if there exist polynomials $1/t(n, 1/\epsilon)$, $s(n, 1/\epsilon)$, $p(n, 1/\epsilon)$ and $g(n, 1/\epsilon)$ such that for every $n$, $f \in C$, $\epsilon > 0$, and every $r_0 \in R$, with probability at least $1 - \epsilon$, a sequence $r_0, r_1, r_2, \ldots$, where $r_i \leftarrow \text{SelNB}[L, t, p, s](f, D, \mathcal{A}, r_{i-1})$ will have $L\text{Perf}_f(r_{g(n,1/\epsilon)}, D) > 1 - \epsilon$.*

As in PAC learning, we say that a concept class $C$ is evolvable if it is evolvable over all distributions by a single evolution algorithm (we emphasize this by saying *distribution-independently* evolvable). Similarly, we say that a class of distribution-function pairs $\mathcal{Z}$ is evolvable if the evolution algorithm is successful for all pairs $(D, f) \in \mathcal{Z}$.

We say that an evolution algorithm $\mathcal{A}$ evolves $C$ over $D$ *monotonically* if with probability at least $1 - \epsilon$, for every $i \leq g(n, 1/\epsilon)$, $L\text{Perf}_f(r_i, D) \geq L\text{Perf}_f(r_0, D)$, where $g(n, 1/\epsilon)$ and $r_0, r_1, r_2, \ldots$ are defined as above. Note that since the evolution algorithm can be started in any representation, this is equivalent to requiring that with probability at least $1 - \epsilon$, $L\text{Perf}_f(r_{i+1}, D) \geq L\text{Perf}_f(r_i, D)$ for every $i$.

## 3 Lower Bounds on Distribution-Independent CSQ Learnability

In this section we demonstrate that conjunctions are not evolvable with Boolean loss (or the equivalent linear loss). We obtain this result by exploiting the equivalence of evolvability with Boolean loss and efficient CSQ learnability. Our technique is based on a combinatorial parameter of a concept class $C$, referred to CSQD that lower bounds the complexity of distribution-independent CSQ learning of $C$. This parameter can be seen as a generalization of the approximation-based strong statistical query dimension given in our earlier work [6] to the distribution-independent setting.

**Definition 3.1** *For a concept class $C$, and $\epsilon, \tau > 0$ we define CSQD$(C, \epsilon, \tau)$ as the smallest number $d$ for which it holds that for every distribution $D$ and function $\psi \in \mathcal{F}_1^\infty$, there exists a set of $d$ functions $G_\psi \subset \mathcal{F}_1^\infty$ and a Boolean function $h_\psi$ such that for every $f \in C$ and distribution $D'$, at least one of the following conditions holds:*

1. *there exists $g \in G_\psi$ such that $|\langle f, g \rangle_{D'} - \langle \psi, g \rangle_D| \geq \tau$ or*

2. $\mathbf{Pr}_{D'}[f(x) \neq h_\psi(x)] \leq \epsilon.$

We now give a simple proof that CSQD$(C, \epsilon, \tau)$ lower bounds the number of correlational statistical queries of tolerance $\tau$ required to learn $C$ distribution-independently to accuracy $\epsilon$. Our proof is based on the proof of the analogous result for the strong SQ dimension [6].

**Theorem 3.2** *If $C$ is learnable by a deterministic CSQ algorithm that uses $q(n, 1/\epsilon)$ queries of tolerance $\tau(n, 1/\epsilon)$ then CSQD$(C, \epsilon, \tau(n, 1/\epsilon)) \leq q(n, 1/\epsilon)$.*

**Proof:** Let $\mathcal{A}$ be the assumed CSQ learning algorithm for $C$. Let $\psi \in \mathcal{F}_1^\infty$ be any function and $D$ be any distribution. The set $G_\psi$ and function $h_\psi$ are constructed as follows. Simulate algorithm $\mathcal{A}$ and for every correlational query $(\phi_i \cdot \ell, \tau)$ add $\phi_i$ to $G_\psi$ and respond with $\langle \psi, \phi_i \rangle_D = \mathbf{E}_D[\phi_i(x) \cdot \psi(x)]$ to the query. Continue the simulation until $\mathcal{A}$ outputs a hypothesis. Let $h_\psi$ be the hypothesis output by $\mathcal{A}$.

First, by the definition of $G_\psi$, $|G_\psi| \leq q(n, 1/\epsilon)$. Now, let $f$ be any function in $C$ and $D'$ be a distribution. If there does not exist $g \in G_\psi$ such that $|\langle f, g \rangle_{D'} - \langle \psi, g \rangle_D| \geq \tau$ (the first condition) then for every correlational query function $\phi_i \in G_\psi$, $\langle \psi, \phi_i \rangle_D$ is within $\tau$ of $\langle f, \phi_i \rangle_{D'}$. Therefore the answers provided by our simulator are valid for the execution of $\mathcal{A}$ when the target function is $f$ and the distribution is $D'$. That is they could have been returned by STAT$(f, D')$ with tolerance $\tau$. Therefore, by the definition of $\mathcal{A}$, the hypothesis $h_\psi$ satisfies $\mathbf{Pr}_{D'}[f(x) \neq h_\psi(x)] \leq \epsilon$ (the second condition). ∎

## 3.1 Conjunctions are not CSQ Learnable Distribution-Independently

We now demonstrate that for a carefully constructed set of distributions and conjunctions, no polynomial-size approximating set satisfying the conditions of Definition 3.1 exists. Let $U$ be the uniform distribution over $X = \{0, 1\}^n$. For a set $S \subseteq [n]$ we denote by $t_S(x)$ a conjunction of the variables with indices in $S$ and by $\chi_S(x)$ the parity function of the variables with indices in $S$. A well-known fact about the Fourier representation of conjunctions (e.g. [12]) is that

$$t_S(x) = -1 + 2^{-|S|+1} \sum_{I \subseteq S} \chi_I(x) \,.$$

To obtain the desired lower bound we note that any pair $(D, g)$ where $g$ is a real-valued function over $\{0, 1\}^n$ and $D$ is a distribution can be viewed as a real-valued function $g'(x) = g(x)D(x)/U(x) = 2^n \cdot g(x)D(x)$. Here and below $D(x)$ refers to the probability density function of $D$. By definition, for every $x$, $g(x)D(x) = g'(x)U(x)$ and therefore for any real-valued function $h$, $\langle h, g \rangle_D = \langle h, g' \rangle_U$. This simple transformation allows us to view distribution-function pairs as functions over the uniform distribution and vice versa.

The basis of our constructions are functions whose Fourier transform equals to the Fourier transform of $t_S(x)$ but with all the Fourier coefficients for non-empty sets of size at most $k/3$ removed. We claim that these functions can be seen as conjunctions over a close-to-uniform distribution.

**Lemma 3.3** *Let $k \geq 6$ be an integer divisible by 3 and let $S \subset [n]$ be any set of size $k$. There exists a function $\theta_S(x)$ and distribution $D_S$ such that for every point $x$, $D_S(x)t_S(x) = U(x)\theta_S(x)$ and in addition*

1. *$\theta_S(x) = \alpha \left(-1 + 2^{-|S|+1} + \sum_{I \subseteq S, |I|>k/3} \chi_I(x)\right)$ for a constant $\alpha \in [2/3, 2]$.*

2. *for every $x$, $D_S(x)/U(x) \in [1/3, 3]$.*

**Proof:** Let $\phi_S(x) = -1 + 2^{-k+1} + \sum_{I \subseteq S, |I|>k/3} \chi_I(x)$, in other words $t_S$ with all the parities for subsets of size $i \in [k/3]$ removed. Note that the total number of parities that were removed from $t_S(x)$ is $\binom{k}{k/3} - 1 < 2^{k-2}$. Therefore for every $x$,

$$|t_S(x) - \phi_S(x)| \leq 2^{-k+1} \left| \sum_{I \subseteq S, |I| \in [k/3]} \chi_I(x) \right| < 2^{-k+1} \cdot 2^{k-2} = 1/2 \,.$$

This implies that for every $x$, $\text{sign}(\phi_S(x)) = \text{sign}(t_S(x))$ and $L_1(\phi_S) = \mathbf{E}_U[|\phi_S(x)|] \in [1/2, 3/2]$. Now let $D_S(x) = U(x) \cdot |\phi_S(x)|/L_1(\phi_S)$ and $\theta_S(x) = \phi_S(x)/L_1(\phi_S)$. This definition implies that $\sum_{x \in X} D_S(x) = \mathbf{E}_U[|\phi_S(x)|]/L_1(\phi_S) = 1$. Hence $D_S(x)$ is a valid probability density function over $\{0, 1\}^n$. Further, $D_S(x)t_S(x) = U(x)\theta_S(x)$. In other words the conjunction $t_S$ over the distribution $D_S$ can be viewed as the function $\theta_S(x)$ over the uniform distribution. Finally, note that $\alpha = 1/L_1(\phi_S) \in [2/3, 2]$ and $D_S(x)/U(x) = |\phi_S(x)|/L_1(\phi_S) \in [1/3, 3]$. ∎

We now establish that the number of monotone conjunctions of $k$ variables such any two conjunctions share at most $k/3$ variables is large.

**Lemma 3.4** *For any integer $k \in [6..n/2]$ divisible by 3, there exists a set $\mathcal{S}_k \subseteq 2^{[n]}$, such that*

- *for every $S \in \mathcal{S}_k$, $|S| = k$;*

- *for every distinct $S, T \in \mathcal{S}_k$, $|S \cap T| \leq k/3$;*

- *$|\mathcal{S}_k| \geq (n/(8k))^{k/3} + 1$.*

**Proof:** There are $\binom{n}{k}$ different size-$k$ subsets of $[n]$ and each subset of size $k$ shares more than $k/3$ elements with at most $\binom{k}{k/3}\binom{n-k/3}{2k/3}$ other subsets of size $k$. Hence by greedily constructing $\mathcal{S}_k$ we will obtain at least

$$\frac{\binom{n}{k}}{\binom{k}{k/3}\binom{n-k/3}{2k/3}} = \frac{n! \cdot (2k/3)!^2 \cdot (k/3)!}{(n-k/3)! \cdot k!^2} \geq \left(\frac{n}{8k}\right)^{k/3} + 1$$

subsets. ∎

We are now ready to show that conjunctions of superconstant size are not CSQ learnable to subconstant accuracy. Let $C_k$ denote the concept class of conjunctions of size at most $k$.

**Theorem 3.5** *If $C_k$ is CSQ learnable to accuracy $\epsilon \leq 2^{-k}/6$ by a deterministic algorithm that uses $q$ queries of tolerance $\tau$ then $q/\tau^2 \geq (\frac{n}{8k})^{k/3}/16$.*

**Proof:** We apply Theorem 3.2 to the assumed CSQ algorithm for $C_k$ and obtain that $\text{CSQD}(C_k, \epsilon, \tau) \leq q$. Let $\psi(x) \equiv \alpha\left(-1 + 2^{-k+1}\right)$ and $D$ be the uniform distribution. By the Definition 3.1, there exists a set $G$ of $q$ functions and a Boolean function $h$ such that for every $f \in C_k$ and distribution $D'$ at least one of the following conditions holds:

1. $\mathbf{Pr}_{D'}[f(x) \neq h(x)] \leq \epsilon$ or

2. there exists $g \in G$ such that $|\langle f, g \rangle_{D'} - \langle \psi, g \rangle_U| \geq \tau$.

Let $\mathcal{S}_k$ be the set given by Lemma 3.4 and $S \in \mathcal{S}_k$. We apply these conditions to $f = t_S$ and distribution $D_S$ defined in Lemma 3.3 to obtain that $\mathbf{Pr}_{D_S}[t_S(x) \neq h(x)] \leq \epsilon$ or there exists $g \in G$ such that $|\langle t_S, g \rangle_{D_S} - \langle \psi, g \rangle_U| \geq \tau$. We first consider the implications of the first condition. By our assumption $\epsilon \leq 2^{-k}/6$. For any two subsets $S, T \in \mathcal{S}_k$, $\mathbf{Pr}_U[t_S \neq t_T] > 2^{-k}$. This implies that if $\mathbf{Pr}_{D_S}[t_S \neq h] \leq \epsilon$ then

$$\mathbf{Pr}_U[t_S \neq h] = \sum_{t_S(x) \neq h(x)} U(x) \leq^{(*)} \sum_{t_S(x) \neq h(x)} 3 \cdot D_S(x) = 3 \cdot \mathbf{Pr}_{D_S}[t_S \neq h] \leq 3\epsilon \leq 2^{-k}/2, \quad (1)$$

where $(*)$ is implied by property 2 in Lemma 3.3. Further, $\mathbf{Pr}_U[t_T \neq h] \geq \mathbf{Pr}_U[t_S \neq t_T] - \mathbf{Pr}_U[t_S \neq h] > 2^{-k}/2$ and hence, by the same argument as equation (1),

$$\mathbf{Pr}_{D_T}[t_T \neq h] \geq \mathbf{Pr}_U[t_T \neq h]/3 > (2^{-k}/2)/3 = 2^{-k}/6 \geq \epsilon .$$

In other words, $h$ can be $\epsilon$ close to at most one conjunction $t_S$ for $S \in \mathcal{S}_k$.

Now consider a subset $S$ for which the second condition holds. By the definition of $D_S$, $\langle t_S, g \rangle_{D_S} = \langle \theta_S, g \rangle_U$ and therefore the second condition is equivalent to

$$|\langle \theta_S - \psi, g \rangle_U| \geq \tau .$$

We observe that $\theta_S - \psi = \alpha 2^{-k+1} \cdot \sum_{I \subseteq S, |I| > k/3} \chi_I$ and hence

$$\tau \leq \left| \langle \alpha 2^{-k+1} \cdot \sum_{I \subseteq S, |I| > k/3} \chi_I, g \rangle_U \right| \leq \alpha 2^{-k+1} \cdot \sum_{I \subseteq S, |I| > k/3} |\langle \chi_I, g \rangle_U|. \quad (2)$$

Equation (2) implies that there exists $I_S \subseteq S$ such that $|I_S| > k/3$ and

$$|\langle \chi_{I_S}, g \rangle_U| \geq \tau \cdot 2^{k-1}/(\alpha 2^k) \geq \tau/(2 \cdot \alpha) \geq \tau/4 .$$

We now need two crucial observations:

1. For distinct $S, T \in \mathcal{S}_k$, $I_S \neq I_T$. This is true since $I_S$ is a subset of size at least $k/3 + 1$ of $S$ and $S$ shares at most $k/3$ elements with $T$ (of which $I_T$ is a subset).

2. For any function $g \in \mathcal{F}_1^\infty$, there exist at most $16/\tau^2$ sets $I$ such that $|\langle \chi_I, g \rangle_U| \geq \tau/4$. This is true since $\langle \chi_I, g \rangle_U$ is simply the Fourier coefficient of $g$ with index $I$ denoted by $\hat{g}(I)$. Parseval's identity states that $\sum_{I \subset [n]} \hat{g}(I)^2 = \|g\|_U^2 \leq 1$ and therefore no more than $16/\tau^2$ Fourier coefficients of $g$ can be larger than $\tau/4$.

Combining these two observations gives that the number of subsets of $\mathcal{S}_k$ for which the second condition holds is at most $16 \cdot q/\tau^2$. By combining this with the fact that the first condition can hold for at most one set in $\mathcal{S}_k$ we obtain that $16 \cdot q/\tau^2 \geq |\mathcal{S}_k| - 1 \geq (\frac{n}{8k})^{k/3}$. ∎

**Remark 3.6** *Theorem 3.5 also applies to CSQ learning by randomized algorithms since a randomized algorithm for the set of conjunction-distribution pairs we consider can be converted to a non-uniform deterministic algorithm via a standard transformation (e.g. [1]).*

**Corollary 3.7** *For any $k = \omega(1)$ and $\epsilon = o(1)$, $C_k$ is not evolvable (distribution-independently) to accuracy $\epsilon$.*

Interestingly, conjunctions are known to be weakly CSQ learnable (distribution-independently). Therefore Corollary 3.7 also implies that traditional boosting algorithms [24, 10] cannot be adapted to CSQ learning (and hence evolvability).

## 4 Evolvability of Halfspaces with Non-Linear Loss Functions

In this section we demonstrate that halfspaces are evolvable distribution-independently for a wide class of loss functions using a polynomial in $n$, $1/\epsilon$ and $1/\gamma$ amount of resources. Here $\gamma$ is the margin of the target halfspace on the domain $X$ of the learning problem. For example, if we set $X = \{-1/\sqrt{n}, 1/\sqrt{n}\}^n$ (or the Boolean hypercube scaled to fit in $B_n$) then all functions that can be represented by a halfspace with integer weights upper-bounded in absolute value by $m$, will have the margin of at least $1/(nm)$. Consequently, our result implies that such functions are evolvable distribution-independently over the Boolean hypercube for any $m$ upper-bounded by a polynomial in $n$. This class of functions includes conjunctions, disjunctions, decision lists of length $O(\log n)$ and majority functions. The mutation algorithm we use is very simple and natural for evolving halfspaces. The only operations it requires are adding $\alpha \cdot x_i$ to the current function for a real $\alpha$ and bounding the value of the function to be in $[-1, 1]$.

A more general way to describe this result is to take the domain to be $B_n$ and define the margin relative to the support of the target distribution. Specifically, let $\text{HS}_\gamma$ denote the set of distribution-function pairs over $B_n$ such that $(D, f) \in \text{HS}_\gamma$, if (and only if) $f$ can be represented by a halfspace with margin $\gamma$ on the support of distribution $D$ (for brevity we use "margin on $D$" to refer to the margin on the support of $D$). For $X \subseteq B_n$, we denote by $\text{HS}_\gamma(X)$ the set of all functions that can be represented by a halfspace with margin $\gamma$ on $X$.

Our proof of evolvability relies on the lemma which proves that for every current hypothesis $\phi \in \mathcal{F}_1^\infty$, there exists an efficiently computable and small neighborhood $N(\phi)$ of $\phi$ such that for every target halfspace $f$ with margin $\gamma$ on distribution $D$, if the fitness of $\phi$ is not $\epsilon$-close to the optimum then there exist $\phi' \in N$ whose fitness is observably higher than the fitness of $\phi$. Following Kanade et al. [14], we refer to such function $N$ as *strictly beneficial* neighborhood function. Strictly beneficial neighborhood function immediately implies monotone evolvability from any starting function [6]. To see this observe that for a mutation algorithm that produces a random member of the strictly beneficial neighborhood, every step of the evolution algorithm will increase performance by an inverse-polynomial amount until it reaches $1 - \epsilon$. Further as it was observed by Kanade et al. [14], it also implies evolvability when the target function is allowed to change gradually, or *drift*.

We first show the existence of a strictly beneficial neighborhood function for halfspaces with the quadratic loss function and then examine the conditions on the loss function that allow a similar argument to go through. For $a \in \mathbb{R}$, define

$$P_1(a) \triangleq \begin{cases} a & |a| \leq 1 \\ \text{sign}(a) & \text{otherwise.} \end{cases}$$

**Theorem 4.1** *For $\phi(x) \in \mathcal{F}_1^\infty$, let*

$$N_\alpha(\phi) = \{P_1(\phi + \alpha' \cdot x_i) \mid i \in [0..n], |\alpha'| = \alpha\} \cup \{\phi\}.$$

*For every halfspace $f$ with margin $\gamma$ on distribution $D$ and every $\epsilon > 0$, there exists $\phi' \in N_\alpha(\phi)$ for which*

$$\|f - \phi'\|_D^2 \leq \max\left\{\|f - \phi\|_D^2 - \alpha^2, \epsilon\right\},$$

*where $\alpha = \frac{\epsilon \gamma}{3\sqrt{n}}$.*

**Proof:** Let $f = \text{sign}(\sum_{i \in [n]} w_i x_i - \theta)$ be the representation of $f$ that has margin $\gamma$ on $D$. The claim holds if $\|f - \phi\|_D^2 \leq \epsilon$. We can therefore assume that $\|f - \phi\|_D^2 > \epsilon$. In particular, since for every $a \in [-2, 2]$, $|a| \geq a^2/2$ we obtain that $\mathbf{E}_D[|f - \phi|] \geq \epsilon/2$.

For every $x$ in the support of $D$, $f(x) - \phi(x)$ has the same sign as $f(x)$ and therefore also the same sign as $\sum_{i \in [n]} w_i x_i - \theta$. Therefore,

$$\mathbf{E}_D\left[(f - \phi)\left(\sum_{i \in [n]} w_i x_i - \theta\right)\right] \geq \gamma \mathbf{E}_D[|f - \phi|] \geq \epsilon \gamma / 2. \tag{3}$$

At the same time, using the Cauchy-Schwartz inequality we can obtain

$$\mathbf{E}_D\left[(f - \phi)\left(\sum_{i \in [n]} w_i x_i - \theta\right)\right] \leq \sum_{i \in [n]} |w_i| \mathbf{E}_D[|(f - \phi) x_i|] + |\theta| \mathbf{E}_D[(f - \phi)] \leq \sqrt{\theta^2 + \sum_{i \in [n]} w_i^2} \cdot \sqrt{\sum_{i \in [0..n]} \mathbf{E}_D[(f - \phi) x_i]^2}$$

$$\leq \sqrt{2} \sqrt{\sum_{i \in [0..n]} \mathbf{E}_D[(f - \phi) x_i]^2}. \tag{4}$$

By combining equations (3) and (4) we obtain that

$$\sum_{i\in[0..n]} \mathbf{E}_D[(f-\phi)x_i]^2 \geq (\epsilon\gamma)^2/8 \ .$$

From here we can conclude that there exists $j \in [0..n]$ such that

$$|\mathbf{E}_D[(f-\phi)x_j]| \geq \epsilon\gamma/\sqrt{8(n+1)} \geq \epsilon\gamma/(3\sqrt{n}). \tag{5}$$

Now we claim that a step in the direction of $x_j$ from $\phi$ will decrease the distance (in $\|\cdot\|_D$ norm) to $f$. Formally,

**Lemma 4.2** *Let $\alpha' = \alpha \cdot \mathtt{sign}(\mathbf{E}_D[(f-\phi)x_j])$, where $\alpha = \frac{\epsilon\gamma}{3\sqrt{n}}$ (as defined in the statement of the theorem). Then*

$$\|f - (\phi + \alpha' \cdot x_j)\|_D^2 \leq \|f - \phi\|_D^2 - \alpha^2 \ .$$

**Proof:**

$$\|f - (\phi + \alpha' \cdot x_j)\|_D^2 = \|f - \phi\|_D^2 + \alpha'^2\|x_j\|_D^2 - 2\langle f - \phi, \alpha' \cdot x_j\rangle_D.$$

To obtain the claim it remains to observe that $\|x_j\|_D^2 \leq 1$ and that

$$2\langle f - \phi, \alpha' \cdot x_j\rangle_D = 2\alpha'\mathbf{E}_D[(f-\phi)x_j] \geq 2\alpha'^2 = 2\alpha^2 \ .$$

∎(Lem. 4.2)

Now let $\phi' = P_1(\phi + \alpha' \cdot x_j)$. If for a point $x$, $\phi'(x) = \phi(x) + \alpha' \cdot x_j$ then clearly $f(x) - \phi'(x) = f(x) - (\phi(x) + \alpha' \cdot x_j)$. Otherwise, if $|\phi(x) + \alpha' \cdot x_j| > 1$ then $\phi'(x) = \mathtt{sign}(\phi(x) + \alpha' \cdot x_j)$ and for any value $f(x) \in \{-1, 1\}$, $|f(x) - \phi'(x)| \leq |f(x) - (\phi(x) + \alpha' \cdot x_j)|$. This implies that

$$\|(f - \phi')\|_D^2 \leq \|f - (\phi + \alpha' \cdot x_j)\|_D^2 \leq \|f - \phi\|_D^2 - \alpha^2.$$

By definition, $\phi' \in N_\alpha(\phi)$ and hence we obtain the claimed result. ∎

We now demonstrate that a similar result can be obtained under several mild conditions on the loss function. In essence, we require that the loss function can be well approximated by a linear function with a slope that is not too close to 0. Formally,

**Definition 4.3** *For positive constants $a, A$ and $B$ we say that a loss function $L : \{-1, 1\} \times [-1, 1] \to \mathbb{R}^+$ is well-behaved with bounds $a, A, B$ if*

1. $L(-1, -1) = L(1, 1) = 0$;

2. $L(1, -1) = L(-1, 1) = 2$;

3. *for $\ell \in \{-1, 1\}$, $L(\ell, z)$ is twice differentiable in [-1,1] (the differentiation is always in the second variable);*

4. *for $\ell \in \{-1, 1\}$, $L'(\ell, \ell) = 0$ and $-\ell \cdot L'(\ell, \ell(1 - z)) \geq A \cdot L(\ell, \ell(1 - z))^a$;*

5. *for $\ell \in \{-1, 1\}$, for every $z \in [-1, 1]$, $L''(\ell, z) \leq B$.*

We remark that condition (2) is for convenience only and can be achieved by scaling any loss function satisfying the other conditions. Condition (4) ensures that the loss function is monotone (that is for all $y, y' \in [-1, 1]$, if $y \leq y'$ then $L(-1, y) \leq L(-1, y')$ and $L(1, y') \leq L(1, y)$) and that it has a non-negligible slope whenever the loss itself is non-negligible. Condition (5) ensures that the linear approximation to $L$ dominates the remainder term in the Taylor series. A simple example of a well-behaved loss function is $L(y, z) = |y - z|^c/2^{c-1}$ for any constant $c \geq 2$. It is also easy to see that any convex combination of well-behaved loss functions is well-behaved. We now prove a generalization of Theorem 4.1 to well-behaved loss functions.

**Theorem 4.4** *Let L be a well-behaved loss function with bounds a, A and B. For $\phi(x) \in \mathcal{F}_1^\infty$, let*

$$N_\alpha(\phi) = \{P_1(\phi + \alpha' \cdot x_i) \mid i \in [0..n], |\alpha'| = \alpha\} \cup \{\phi\}.$$

*For every halfspace $f$ with margin $\gamma$ on distribution $D$ and every $\epsilon > 0$, there exists $\phi' \in N_\alpha(\phi)$ for which*

$$\mathbf{E}_D[L(f, \phi')] \leq \max\{\mathbf{E}_D[L(f, \phi)] - \alpha^2 \cdot B/2, \epsilon\},$$

*where $\alpha = A \cdot \gamma \cdot \epsilon^{a+1}/(B \cdot 2^{a+3}\sqrt{n})$.*

**Proof:** As before, we can assume that $\mathbf{E}_D[L(f, \phi)] > \epsilon$. In particular, $\mathbf{Pr}_D[L(f, \phi) \geq \epsilon/2] \geq \epsilon/4$. Then, by property (4) of well-behaved loss-functions, $\mathbf{Pr}_D[|L'(f, \phi)| \geq A(\epsilon/2)^a] \geq \epsilon/4$. This implies that

$$\mathbf{E}_D[|L'(f, \phi)|] \geq \epsilon/4 \cdot A \cdot (\epsilon/2)^a = A \cdot \epsilon^{a+1}/2^{a+2}.$$

By monotonicity of $L$ (or property (4)), for every $x$ in the support of $D$, $-L'(f(x), \phi(x))$ has the same sign as $f(x)$ and therefore also the same sign as $\sum_{i \in [n]} w_i x_i - \theta$. This gives

$$\mathbf{E}_D\left[-L'(f, \phi)\left(\sum_{i \in [n]} w_i x_i - \theta\right)\right] \geq \gamma \mathbf{E}_D[|L'(f, \phi)|] \geq A \cdot \gamma \epsilon^{a+1}/2^{a+2}. \tag{6}$$

In addition, as in equation (4), we have

$$\mathbf{E}_D\left[-L'(f, \phi)\left(\sum_{i \in [n]} w_i x_i - \theta\right)\right] \leq \sqrt{2}\sqrt{\sum_{i \in [0..n]} \mathbf{E}_D[L'(f, \phi) \cdot x_i]^2}. \tag{7}$$

By combining equations (6) and (7) we obtain that

$$\sum_{i \in [0..n]} \mathbf{E}_D[L'(f, \phi)x_i]^2 \geq A^2 \cdot \gamma^2 \epsilon^{2a+2}/2^{2a+5}.$$

From here we can conclude that there exists $j \in [0..n]$ such that

$$|\mathbf{E}_D[L'(f, \phi)x_j]| \geq A \cdot \gamma \cdot \epsilon^{a+1}/(2^{a+2}\sqrt{2n+2}) \geq A \cdot \gamma \cdot \epsilon^{a+1}/(2^{a+3}\sqrt{n}). \tag{8}$$

We denote the right side of this inequality by $\rho$.

To finish the proof we prove an analogue of Lemma 4.2 saying that a step in the direction of $x_j$ from $\phi$ will decrease the loss. Formally,

**Lemma 4.5** *Let $\alpha' = -\alpha \cdot \mathtt{sign}(\mathbf{E}_D[L'(f, \phi)x_j])$, and $\phi' = P_1(\phi + \alpha' \cdot x_j)$, where $\alpha = \rho/B$ (as defined in the statement of the theorem). Then*

$$\mathbf{E}_D[L(f, \phi')] \leq \mathbf{E}_D[L(f, \phi)] - \alpha^2 \cdot B/2.$$

**Proof:** Let $x \in X$ be any point. Assume that $f(x) = -1$. For convenience we extend the loss function $L(-1, z)$ to values $z \in [-2, -1)$ by setting $L(-1, z) = L(-1, -2 - z)$ (that is by making the loss symmetric around $-1$). By the properties of the loss function, $L(-1, -1) = 0$, $L'(-1, -1) = 0$ and for $z \in [-2, -1)$, $L''(-1, z) = L''(-1, -2 - z)$. This implies that the extended $L$ is twice differentiable in $[-2, 1]$ and $L''(-1, z) \leq B$ for every $z \in [-2, 1]$. We first assume that $\phi + \alpha' \cdot x_j \in [-2, 1]$. $L$ is twice differentiable and therefore Taylor's theorem gives

$$L(-1, \phi(x) + \alpha' \cdot x_j) - L(-1, \phi(x)) = \alpha' \cdot x_j \cdot L'(-1, \phi(x)) + (\alpha' \cdot x_j)^2 \cdot L''(-1, \zeta)/2,$$

where $\zeta \in [\phi(x), \phi(x) + \alpha' \cdot x_j] \subseteq [-2, 1]$. Also note that in this case, $L(-1, \phi(x) + \alpha' \cdot x_j) \geq L(-1, \phi'(x))$. This means that

$$L(-1, \phi'(x)) - L(-1, \phi(x)) \leq \alpha' \cdot x_j \cdot L'(-1, \phi(x)) + \alpha^2 \cdot B/2, \tag{9}$$

Now if $\phi + \alpha' \cdot x_j > 1$ then $\phi'(x) = 1$ and $\alpha' \cdot x_j > 1 - \phi(x) > 0$. Then $\alpha' \cdot x_j \cdot L'(-1, \phi(x)) \geq (1 - \phi(x)) \cdot L'(-1, \phi(x))$ (as $L'(-1, \phi(x)) > 0$). Hence,

$$L(-1, \phi'(x)) - L(-1, \phi(x)) = (1-\phi(x)) \cdot x_j \cdot L'(-1, \phi(x)) + ((1-\phi(x)) \cdot x_j)^2 \cdot L''(-1, \zeta)/2 \leq \alpha' \cdot x_j \cdot L'(-1, \phi(x)) + \alpha^2 \cdot B/2, \quad (10)$$

where $\zeta \in [\phi(x), 1]$. By treating the case when $f(x) = 1$ symmetrically and combining equations (9) and (10) we will obtain that for every $x$,

$$L(f(x), \phi'(x)) - L(f(x), \phi(x)) \leq \alpha' \cdot x_j \cdot L'(f(x), \phi(x)) + \alpha^2 \cdot B/2.$$

This immediately implies that

$$\mathbf{E}_D[L(f(x), \phi'(x))] - \mathbf{E}_D[L(f(x), \phi(x))] \leq \alpha' \mathbf{E}_D[x_j \cdot L'(f(x), \phi(x))] + \alpha^2 \cdot B/2 \leq -\alpha\rho + \alpha^2 \cdot B/2 = \alpha^2 \cdot B/2 \ .$$

∎(Lem. 4.5)

To finish the proof we observe that $\phi'(x) \in N_\alpha(\phi)$. ∎

As we have mentioned, a simple corollary of Theorem 4.4 is distribution-independent evolvability of large margin halfspaces with any well-behaved loss function.

**Theorem 4.6** *For every well-behaved loss function $L$ and $\gamma \geq 1/q(n)$ for some polynomial $q(\cdot)$, $\mathtt{HS}_\gamma$ over $B_n$ is monotonically evolvable with $L$.*

We make two remarks regarding these theorems.

**Remark 4.7** *In both Theorems 4.1 and 4.4 it is not necessary to know the exact value of $\alpha$ to create a strictly beneficial neighborhood. It is easy to see from the analysis that the bound holds for every $\alpha_0 < \max_{j \in [0..n]}\{|\mathbf{E}_D[L'(f, \phi)x_j]|\}$. Therefore by including in the neighborhood steps for all values of $\alpha_0 = 2^{-t}$ for $t \in [n]$, the neighborhood will include a function with at least $1/4$ of the improvement that can be achieved when a bound on $\alpha$ is known in advance.*

**Remark 4.8** *Theorem 4.4 does not require the loss function to be the same for all $x$ as long as for every point $x$, the loss-function $L_x$ is well-behaved with the same bounds $a, A, B$. Similarly the loss function does not need to stay the same between generations and can change arbitrarily as long as it is well-behaved with the same bounds $a, A, B$*

A number of popular machine learning algorithms work by embedding the data points in a different Euclidean space (most commonly by using a kernel) and then applying a learning algorithm for halfspaces, such as SVM. This method is also used in a number of theoretical algorithms such as the DNF learning algorithm based on the polynomial threshold function representation of Klivans and Servedio [18]. As expected, this technique can be easily translated to the evolvability framework and then used together with our result. Formally, let $C$ and $C'$ be concept classes over the domains $X$ and $X'$, respectively. The concept $C$ over $X$ is said to be embeddable as $C'$ over $X'$ if there exists a function $\Phi : X \to X'$ such that for every $f \in C$, there exists $g \in C'$ such that for every $x \in X$, $g(\Phi(x)) = f(x)$. We also say that the embedding is efficient if $\Phi(x)$ is computable efficiently, that is in time polynomial in the dimension of $x$ (or description length in general). Embeddability of concept classes into large-margin halfspaces has been studied in a number of works initiated by Forster [8] and Forster et al. [9] (see [20, 25, 19] for some recent results). The inverse of the optimal margin is referred to as the *margin complexity* of a concept class [20]. Besides its importance to machine learning, it has several connections to fundamental quantities in communication complexity [11, 25, 19]. We cannot invoke this measure directly to upper-bound the complexity of using our evolution algorithm since margin complexity disregards the computational complexity of the embedding function. But given an efficient embedding function the application of our evolution algorithm becomes straightforward.

**Corollary 4.9** *Let $C$ be a concept class over domain $X$, $X' \subseteq B_n$ and $\gamma > 1/q(n)$ for some polynomial $q(\cdot)$. If there exists an efficiently computable embedding of $C$ over $X$ to $\mathtt{HS}_\gamma(X')$ over $X'$, then $C$ is evolvable monotonically with any well-behaved loss function.*